\documentclass[letterpaper]{article}
\pdfoutput=1 
\usepackage{aaai}
\usepackage{times}
\usepackage{helvet}
\usepackage{courier}
\usepackage{subfigure}

\usepackage[dvips]{graphicx}
\usepackage[usenames,dvipsnames]{color}

\begin{document}

\title{Between conjecture and memento: \\shaping a collective emotional perception of the future}
\author{Alberto Pepe \\
Graduate School of Information \\
University of Califonia, Los Angeles \\
\texttt{apepe@ucla.edu}
\And Johan Bollen \\
Digital Library Research and Prototyping \\
Los Alamos National Laboratory\\
\texttt{jbollen@lanl.gov}
}
\maketitle
\begin{abstract}
\begin{quote}
Large scale surveys of public mood are costly and often impractical to perform. However, the web is awash with material indicative of public mood  such as blogs, emails, and web queries. Inexpensive content analysis on such extensive corpora can be used to assess public mood fluctuations. The work presented here is concerned with the analysis of the public mood towards the future. Using an extension of the Profile of Mood States questionnaire, we have extracted mood indicators from 10,741 emails submitted in 2006 to futureme.org, a web service that allows its users to send themselves emails to be delivered at a later date. Our results indicate long-term optimism toward the future, but medium-term apprehension and confusion.
\end{quote}
\end{abstract}

\section{Introduction and Scope}
The ongoing popularization and growing availability of the Internet has led, in recent years, to a sharp increase in the amount of personal, social and emotional information that is publicly and openly shared. Manifestations of mood can be found in personal blogs, online diaries, social networking websites, message boards and chat rooms. Many of these online services offer ad hoc tools to let their users annotate their writings or webpages with a description of their current mood. For example, popular services such as Livejournal ({\tt http://www.livejournal.com/}), an online blogging software and community, and myspace ({\tt http://www.myspace.com/}), a social networking website, allow users to add a mood indicator to their blog entries and their personal profiles, respectively.     

Besides user-contributed mood descriptions, online textual content of various kinds can be aggregated and analyzed to infer public mood levels. Natural language processing, text mining and machine learning can be employed to assess the mood and opinions of groups of users in an automated fashion. Such analytic methods, often referred to as {\it sentiment analysis} \cite{sentiment:nasukawa2003}, bypass the costs of repeated, large-scale public surveys and may yield accurate results when used on extensive corpora. There exists a considerable amount of experimentation in this field. Researchers at the Information and Language Processing Systems group at the Informatics Institute of the University of Amsterdam have used blog data to deduct trends and seasonality \cite{decomp:balog2006}, forecast mood \cite{capture:mishne2006}, and predict movie sales \cite{movie:mishne2006}. Investigating a similar online corpus, \citeauthor{happy:rada2006} (\citeyear{happy:rada2006}) have performed a ``linguistic ethnography'' to find out where happiness lies in everyday lives. Some similar analytical tools function entirely on the web: We Feel Fine ({\tt http://www.wefeelfine.org/}) constantly harvests blog posts for occurrences of the phrases "I feel" and "I am feeling" and matches it to a pre-identified database of feelings; similarly, Moodviews ({\tt http://moodviews.com}) constantly tracks a stream of Livejournal weblogs that are user-annotated with pre-defined moods. The results generated via the analysis of such collective mood aggregators are compelling and indicate that accurate public mood indicators can be extracted from online materials. However, such indicators are limited to near-present observations due to their source material, i.e.~mood indicators extracted from presently available material pertain to past and present moods. They need to be statistically extrapolated to assess future public mood or sentiment.

In this paper, we investigate the content of emails submitted by anonymous users to the futureme web site ({\tt http://www.futureme.org}). The futureme service allows users to compose email messages to themselves for future delivery; emails are held on the server for a certain time (from 1 month up to 30 years in the future) and then sent to the specified recipients at the specified dates. The content of futureme emails is thus explicitly directed towards a future date. In fact, the most common use of this service is for users to address their future selves in the form of a personal memento or conjecture about their future, e.g. "I am sure your AAAI paper was accepted for publication and you are now happily married." An analysis of the mood of these emails will thus yield a direct, rather than extrapolated, assessment of public mood toward the future.  

Futureme currently hosts nearly 500,000 emails to the future. Although a user might send more than one email, it is reasonable to estimate that the total population of the futureme community is in the range of few hundred thousands. Its policy of limitless access and the breadth of attention that futureme has recently received from the press \cite{press1,press2} has contributed to highly diversify its user community. Nevertheless, the findings presented in this paper are based on a sample of the population that is limited to the community of futureme users, i.e.~those that submitted their emails to the futureme website.

Although the size and characteristics of the futureme community are unknown due its focus on user anonymity, it clearly abides by some of the key criteria required by the "wisdom of crowd" phenomena as observed in finance and prediction markets \cite{wisdom:surowiecki004}. In particular, the futureme crowd manifests: a) independence (users are not influenced by other users while composing their emails), b) decentralization (futureme is an online, thus distributed, community), and finally c) diversity of opinion (users write privately to themselves). 

Instead of relying on user-provided mood annotations and {\it ad hoc} models of mood states, we used a popular psychometric instrument to extract mood indicators from the body of futureme emails submitted in 2006, namely an expanded version of the Profile of Mood States questionnaire (POMS) \cite{manual:mcnair1971}. Mood indicators were thus extracted from 10,741 emails submitted to futureme in 2006 and aggregated to query the ``wise'' futureme crowd to compute collective sentiment towards forthcoming years.

The remainder of the paper is organized as follows. In the next section we discuss the data collection process, the content of the data and some statistical facts about the data employed in the study. In the following section, we introduce the methodology, in particular with respect to the adaptation of the POMS test to extensive email corpora. In the last two sections, we present and discuss our results.       

\section{Data collection and processing.}

The data used in the study was obtained from the archives of futureme, a web service that allows its users to send themselves emails to be delivered at a later date, up to 30 years in the future. Upon submission of the email, users can decide whether to make the content of their email publicly available. Subsequently, futureme publishes the following information for public emails without revealing the user's identity: a) date of email dispatch (when the message was written), b) textual content of email and c) intended date of delivery (when the message is due to arrive).

\begin{figure}[htbp]
\begin{center}
\includegraphics[width=0.47\textwidth]{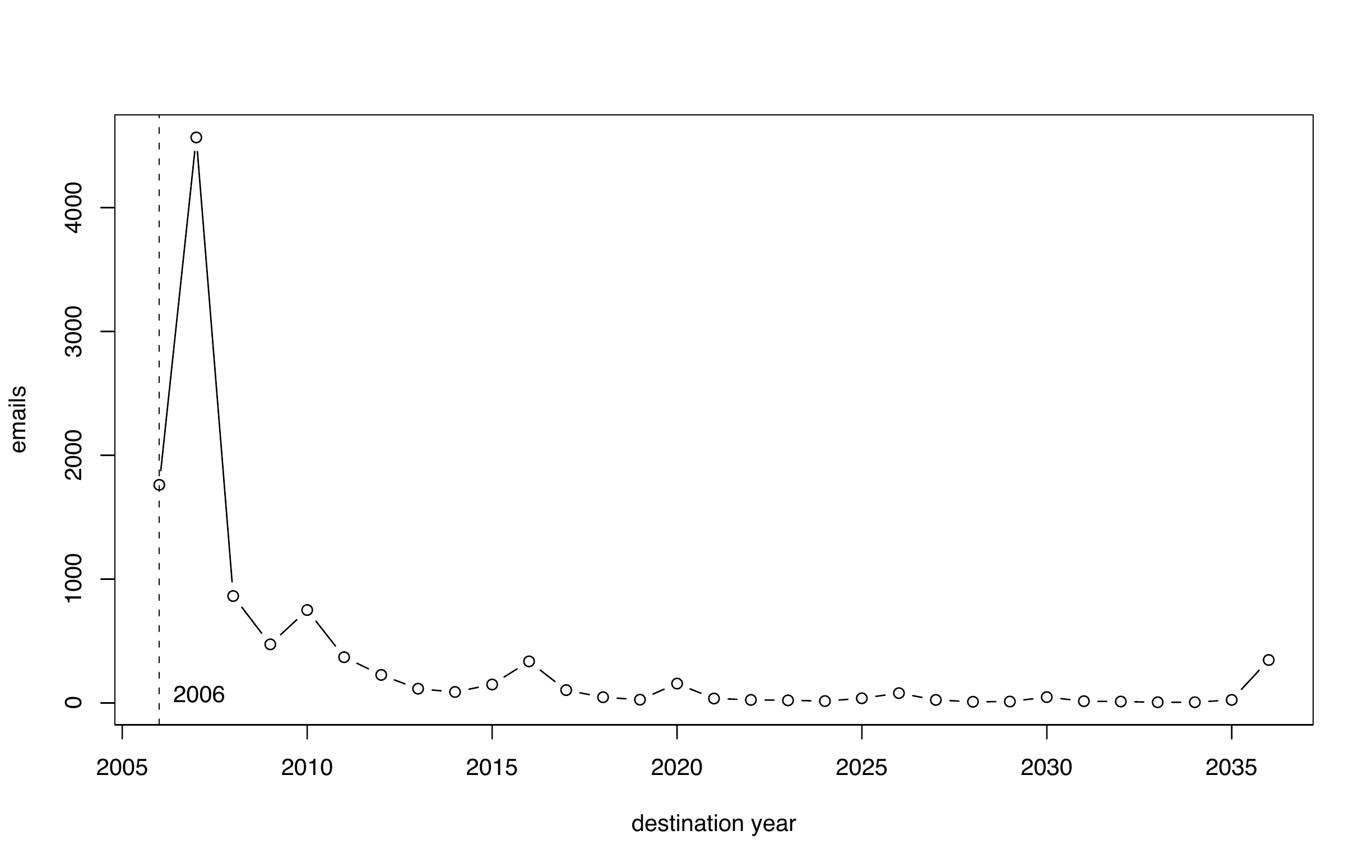}
\caption{Email delivery date distribution for the period 2006 to 2036}
\label{delivery}
\end{center}
\end{figure}

Futureme has experienced significant growth in the 2004 to 2007 period. Whereas in the early years of the service (2002), no more than 500 to 1000 public messages were sent at a yearly basis, this number peaked at about 70,000 in 2006. At the time this analysis was conducted, complete 2007 data was not yet available although extrapolation indicates a continued yet less hyperexponential growth of the service. For this reason, only messages submitted in 2006 were selected for further analysis. All public emails written between January 1 and December 31, 2006 were harvested from the futureme archive and stored in a format suitable for textual analysis. A total of 10,741 emails were collected for the year 2006, with delivery dates ranging from 2006 to 2036. The distributions of emails per destination year as well as the average lag between origin and destination year are shown in Figure \ref{delivery}. 

The graph shows a sharp decrease in the number of emails sent to the distant future. It is very likely that this finding reflects public pessimism about the durability of present-day email addresses, as well as services and technology. Yet, a basic word occurrence analysis of the email corpus under study reveals that the aggregated email content is overall optimistic. Table 1 displays the 20 most frequent terms used in the emails written in 2006. 
\begin{footnotesize}
\begin{table}[h]
\label{wordcount}
\begin{center}
\begin{tabular}{ll|ll}
\textbf{count}&\textbf{word}&\textbf{count}&\textbf{word}\\
\hline\hline
21373&dear&8074&right\\
21318&hope&7931&future\\
15589&love&7184&really\\
12747&know&6625&things\\
11438&life&6103&better\\
10960&time&5974&think\\
9075&happy&5299&email\\
8669&remember&5290&past\\
8643&good&4880&birthday\\
8401&year&4679&today\\
\end{tabular}
\end{center}
\caption{Twenty most recurrent terms in emails composed in 2006, sorted by occurrence}
\end{table}
\end{footnotesize}

Using a combination of manual and automated procedures, we identified and removed emails written (a) in languages other than English and (b) in unknown character encodings. Before proceeding with the content analysis, we visually inspected the email corpus to qualitatively determine whether we could detect similarities or clear patterns in the the nature of the messages. We could identify two broad categories:
\begin{itemize}
	\item {\bf conjectures}: forecasts and predictions about the future, especially with respect to personal life aspects
	\item {\bf mementos}: purposeful reminders and recollections about personal events that might have been forgotten with time  
\end{itemize}

As the study presented here involves a large scale mood analysis, the dichotomy presented above raised an important issue: {\it does an email sent to the future convey the mood of the user at the time of writing, or the expected perception of mood toward the delivery date?} Clearly, emails to the future may contain conjectures or mementos and thus convey mood states relating to either the email composition or delivery date. The distinction between a ``conjecture'' and a ``memento'' email is sometimes too subtle to be detected via manual inspection, let alone via automated techniques. To overcome this obstacle, we dropped this dichotomy in favor of a more holistic approach. In particular, we treated the emails as ``confessional'' time capsules: messages whose content does not necessarily reflect present nor future mood states, but rather a combination of the two. The results of the study presented here must therefore be interpreted in this light. The mood of the emails is intended as the user's present mood state toward the future, similar to how the yield curve in economics reflects present expectations of future yields \cite{rebonato}.

\section{Methodology}
There exist numerous psychometric instruments to assess individual mood states and monitor their fluctuations over time. The Profile of Mood States (POMS) questionnaire \cite{manual:mcnair1971} in particular has enjoyed widespread adoption. POMS measures six dimensions of mood, namely tension-anxiety, depression-dejection, anger-hostility, vigor-activity, fatigue-inertia, and confusion-bewilderment. It does so by having respondents indicate on a five-point intensity scale how well each one of the 65 POMS adjectives describes their present mood state. The respondent's ratings for each mood adjective are then transformed by means of a scoring key to a 6-dimensional mood vector. For example, the user-indicated ratings for terms "angry", "peeved", "grouchy", "spiteful", and "annoyed" all contribute to the respondent's score on the POMS "anger" scale. The POMS is an easy-to-use, low-cost instrument whose factor-analytical structure has been repeatedly validated and recreated \cite{factor:norcross2006}. It has been used in hundreds of studies since its inception \cite{profil:mcnair2003} and has been normed for a wide variety of populations.

To reduce the time and effort on the part of human subjects to complete the POMS, modifications have been proposed that have a reduced number of mood adjectives \cite{shorten1,shorten2}. For the purposes of this study, however, we seek to modify the POMS to apply to open-ended textual material, i.e.~the futureme collection of emails. Since the vocabulary of the futureme emails cannot be controlled, the POMS was then expanded beyond its original set of 65 mood adjectives to match the numerous synonyms and wordings that users may choose to express their mood. For example, whereas the original POMS would require human subjects to rate the term "peeved", the author of a futureme email could have used the term "mad". The occurrence of the term "mad" nevertheless serves as an indication of the user's anger level. It was therefore mapped to the POMS mood adjective "angry".

We extended the intial set of POMS's 65 mood adjectives by 793 synonyms extracted from the Princeton University's WordNet (version 3.0) and Roget's New Millennium Thesaurus (First Edition). For example, the original POMS term {\it discouraged} was assigned a set of extended terms as follows:
\begin{description}
	\item {\it discouraged}, beat down, caved in, crestfallen, daunted, deterred, dispirited, downbeat, downcast, glum, lost momentum, pessimistic, put off
\end{description}

As such, each original POMS mood adjective, referred to as "main", was now associated with a set of synonyms which we refer to as "extended". We will refer to the resulting extended version of the POMS test, as POMS-l. All POMS-l mood adjectives, both main and extended, were stemmed to a common lexical root using the Porter stemmer \cite{algori:porter1980} so that they would match different versions of the same term occurring in a futureme emails. For example, the email term "angrily" whose Porter stem is "angri" would match the POMS adjective "angry" since the latter would also be Porter-stemmed to their common root "angri". 

To assess the mood of a futureme email, its content was matched against the POMS-l's main and its associated extended set of mood adjectives. If an email term matched either a main term or its extended set of mood adjectives this lead to the corresponding main term's score to be incremented by 1. For example, if an email contained the term "daunted", this term match would occur within the extended set for the main term "discouraged". Consequently, the rate of the POMS-l mood adjective "discouraged" would be incremented by 1. Following this procedure, the content of a particular futureme email could be mapped to a set of scores for each of the original POMS mood adjectives, as if a human subject had individually rated each of them. The resulting scores were then converted to a six-dimensional mood vector. This vector was normalized to unit-length.

As a result, each email was mapped to a normalized six-dimensional POMS mood vector representing its level of tension, depression, anger, vigor, fatigue and confusion. These mood vectors were grouped according to the delivery date of the original email, resulting in a set of mood state vectors for each year between 2006 and 2036. A two-sided Kolmogorov-Smirnov test of significance was then conducted between the sets of mood vectors for each pair of years between 2006 and 2036. Statistically significant mood changes between any two particular years ($p<0.05$) could thus be detected.

The trend lines for each of the mood dimensions were plotted over time. Mean mood scores were converted to z-scores according to the time-series mean and standard deviation to establish an equal scaling for all mood dimension trend lines and to be able to interpret deviations from the temporal mean. A 2-degree polynomial smoothing was applied to the raw data to elucidate long-term trends. 

\section{Results}
All measured dimensions indicated mood fluctuations over time, however, statistically significant differences at the significance level ($p<0.05$) were found in the trend lines of only two scales, namely depression and vigor.

\begin{figure}[h!]	
	\begin{center}
    \mbox{
\subfigure[Fatigue]{\includegraphics[width=0.4\textwidth]{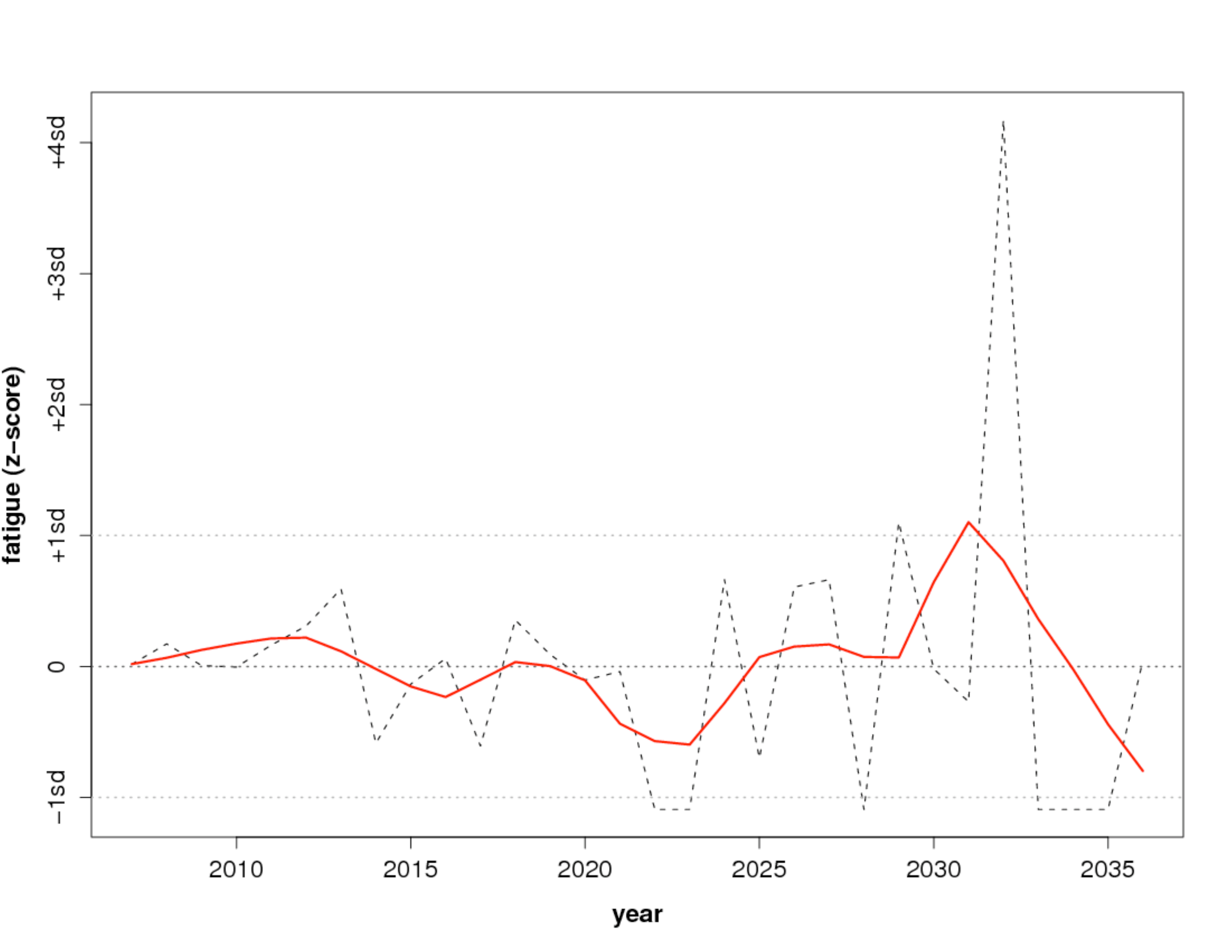}}\quad
}
    \mbox{
\subfigure[Tension]{\includegraphics[width=0.4\textwidth]{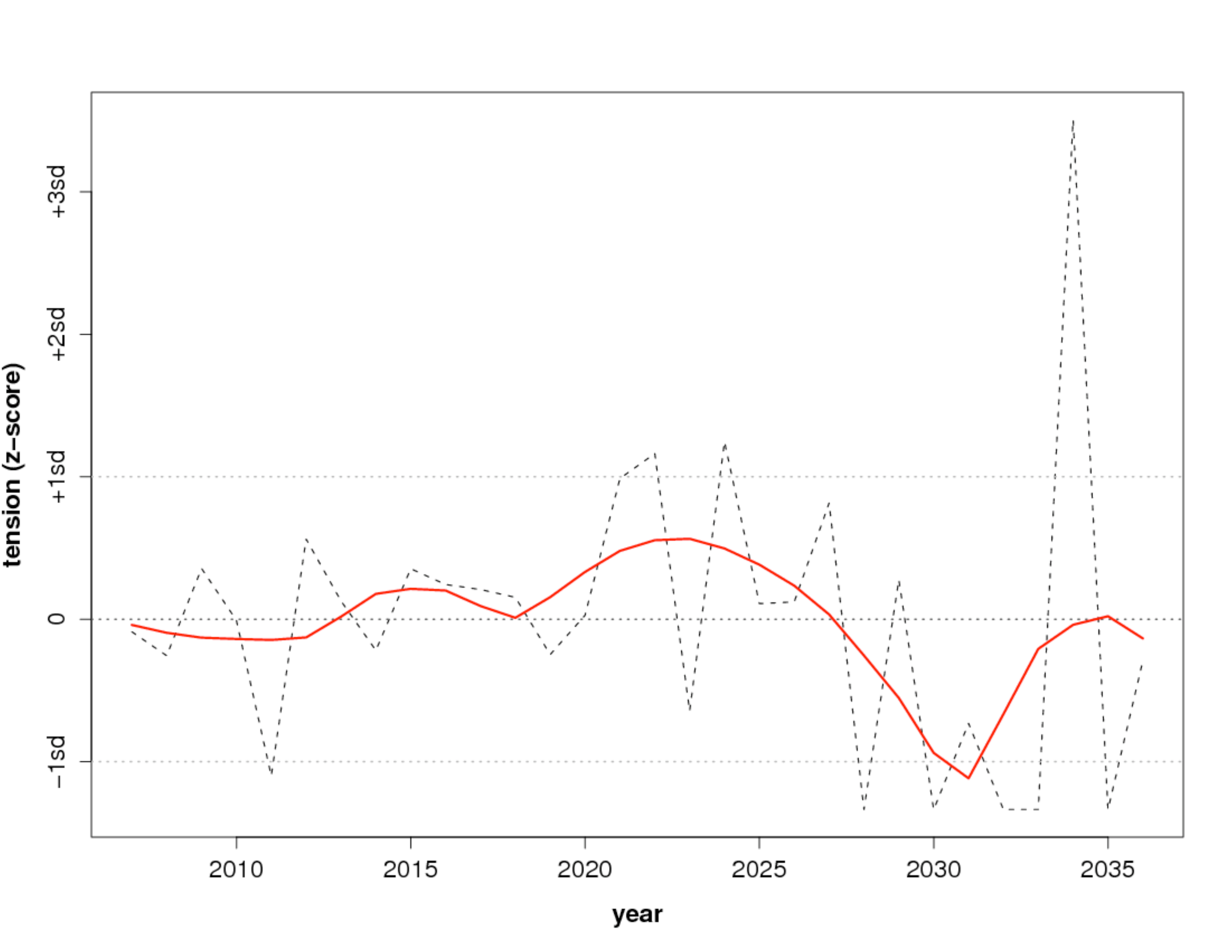}}\quad
}	
    \mbox{
\subfigure[Anger]{\includegraphics[width=0.4\textwidth]{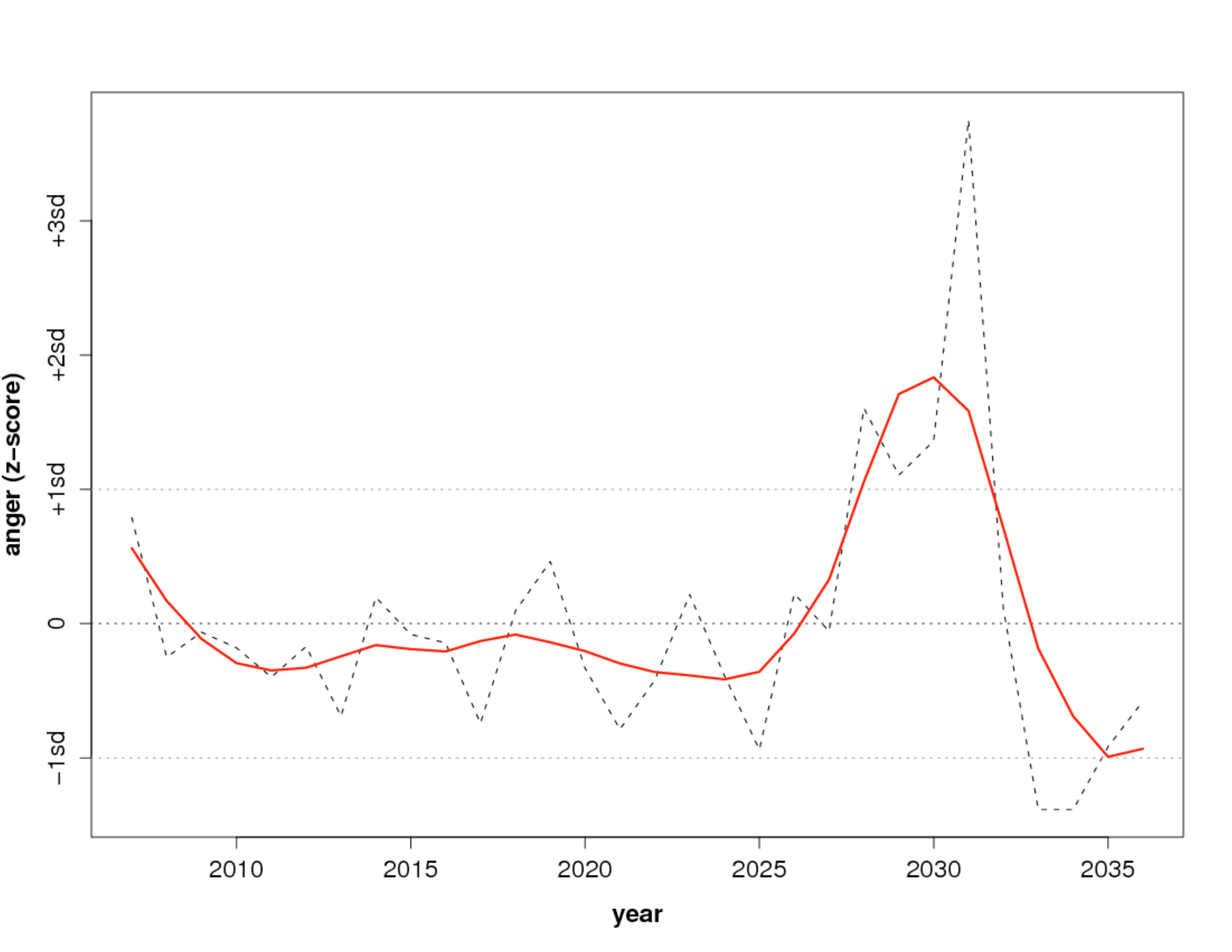}}\quad
}	
\caption{\label{fig:three}Measured fluctuation of mood indicators {\it fatigue}, {\it tension} and {\it anger}.}
	\end{center}
\end{figure}

{\bf Fatigue, tension and anger}. These three mood dimensions do not display major fluctuations over time (Figure \ref{fig:three}). Some spikes can be observed in all three datasets, but since they are not statistically significant, they have been excluded from present discussion. 

\begin{figure}[htbp]
\begin{center}
\includegraphics[width=0.45\textwidth]{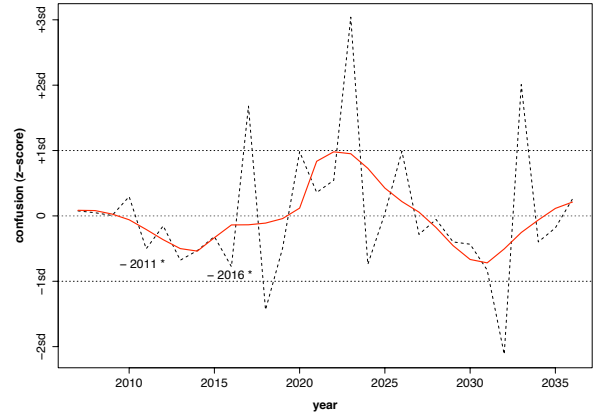}
\caption{Measured fluctuation of mood indicator {\it confusion} (* $p<0.1$)}
\label{confusion}
\end{center}
\end{figure}

{\bf Confusion}. The confusion trend lines (Figure \ref{confusion}) showed only two marginally significant differences in the near term. In particular, confusion scores trended downward between 2009 and 2011 (p=0.088), further confirmed by an overlapping decrease between 2009 and 2016 (p=0.055).

\begin{figure}[htbp]
\begin{center}
\includegraphics[width=0.45\textwidth]{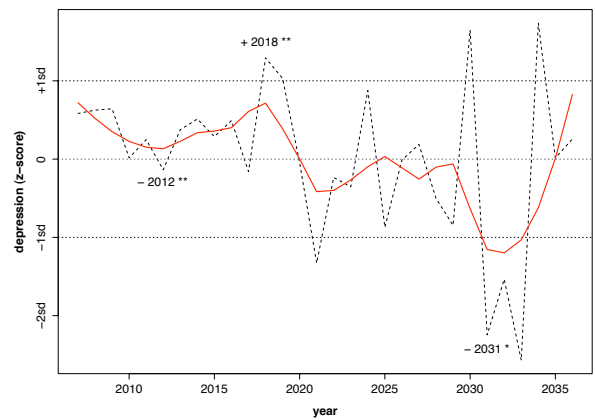}
\caption{Measured fluctuation of mood indicator {\it depression} (* $p<0.1$, ** $p<0.05$)}
\label{depression}
\end{center}
\end{figure}

{\bf Depression}. We observed a statistically significant decrease in depression (Figure \ref{depression}) for the near future, namely the period 2007 to 2012 (p=0.035), followed by a sharp uptick between 2012 and 2018 (p=0.025). Afterward, the depression profile trends downward, however the observed
decline between 2018 and 2031 is only marginally significant (p=0.090). 

\begin{figure}[htbp]
\begin{center}
\includegraphics[width=0.45\textwidth]{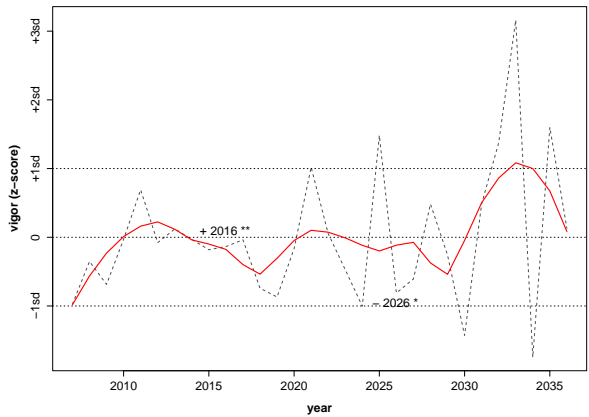}
\caption{Measured fluctuation of mood indicator {\it vigor} (* $p<0.1$, ** $p<0.05$)}
\label{vigor}
\end{center}
\end{figure}

{\bf Vigor}. Vigor (Figure \ref{vigor}) follows, as expected, an inverse pattern to depression. A highly significant increase in vigor scores was observed
between the period 2007 to 2016 (p=0.0042) and a marginally significant decrease between 2016 and 2026 (p=0.085). 

\section{Conclusions}
 
What have we learned from the analysis of a set of personal emails intended to be sent to the future? No statistically significant mood trends have been detected for the fatigue, tension and anger dimensions of the POMS. This could be attributed to either the failure of the POMS-l to detect such changes or to the stability of these mood indicators over time. It could in fact be argued that mood dimensions such as "fatigue" are not appropriate indicators of public mood toward the future. However, it is nevertheless conceivable that levels of tension and anger can change in the public mood towards the future, but our results did not provide reliable indications of changes in these mood dimensions.

The POMS-l did however detect statistically and marginally significant differences among the "confusion", "depression" and "vigor" mood scores of emails directed at specific future dates. The significance values corroborate the generated trend lines shown in Figures \ref{confusion}, \ref{depression} and \ref{vigor}. Marginally significant statistical differences were found in the "confusion" dimension over time. These differences suggest that confusion trends downward in the short time (2009-2011) and increases in the near long term (2009-2016). This fits our intuitive understanding that the near term future is more predictable and therefore less confusing than the long term future. However, we warn that the observed differences are only marginally statistically significant ($p<0.1$). Statistically significant differences were observed for the depression and vigor indicators over time. In particular, depression trends downward for the near future and back up for the 4 year period between (2012 and 2018). The following decrease in depression was only marginally statistically significant, but the overall pattern seems to be one where the futureme community in the mean is optimistic for the short term but less so in the medium term particularly the period between 2012 and 2018. The "vigor" indicator seems to follow an inverse pattern where vigor increases in the short term but decreases in the long term.

The public mood of 30,000 futureme users thus indicates a mixed picture. The collective mood of this user crowd towards the near term future is optimistic, whereas the long term picture is one that seems to indicate a moderate increase in "depression" and a moderate decrease in "vigor". This effect could be tied to the intrinsically personal and intimate nature of the futureme messages. Whereas individuals feel in control and optimistic about their personal prospects in the near future, they may experience apprehension over their own long-term prospects and the world, in general.

These results point to the possibility of extending a tried and true psychometric instrument such as the POMS to assess the public mood from plubic collections of messages directed towards future dates. More efforts need to be dedicated to further validating the proposed POMS-l and increasing its sensitivity. This would however necessitate either the availability of more extensive collections such as the entire futureme body of messages, or a validation effort involving traditional psychometric evaluation measures. Another interesting venue of research would be the validation of our results to other "wisdom of crowds" indicators such as the yield curve in economics and large-scale public surveys.

\section{ Acknowledgments}
We would like to thank Jay Patrikios and Matt Sly for letting us harvest and use the futureme.org public email archive. Also, many thanks to Jean-Fran{\c c}ois Blanchette and Mark Hansen of University of California, Los Angeles for their advice and assistance throughout the early phase of this work.

\bibliographystyle{aaai}

\end{document}